\pdfoutput=1
\documentclass[11pt]{article}
\usepackage{color}
\usepackage[normalem]{ulem}
\usepackage{colortbl}
 \usepackage{booktabs}
\usepackage{multirow}
\usepackage{emnlp2021}
\usepackage{multirow,hhline,graphicx,array}
\usepackage{times}
\usepackage{latexsym}

\usepackage[T2A,T1]{fontenc}
\usepackage[utf8]{inputenc}
\usepackage[russian,english]{babel}
\usepackage{microtype}
\newcommand\blfootnote[1]{%
  \begingroup
  \renewcommand\thefootnote{}\footnote{#1}%
  \addtocounter{footnote}{-1}%
  \endgroup
}

\definecolor{tilde}{RGB}{121, 42, 52}

\title{Dynamic Terminology Integration for COVID-19 and other Emerging Domains}

\author{Toms Bergmanis$^\dagger$$^{\ddagger}$ \and Mārcis Pinnis$^\dagger$$^{\ddagger}$  \\ \\
  $^\dagger$Tilde / Vienības gatve 75A, Riga, Latvia \\
  $^\ddagger$Faculty of Computing, University of Latvia / Raiņa bulv.  19, Riga, Latvia\\ 
  {\tt \{firstname.lastname\}@tilde.lv}
}

\begin{document}
\maketitle
\begin{abstract}
The majority of language domains require prudent use of terminology to ensure  clarity and adequacy of information conveyed.
While the correct use of terminology for some languages and domains can be achieved by adapting general-purpose MT systems on large volumes of in-domain parallel data, such quantities of domain-specific data are seldom available for less-resourced languages and niche domains.  Furthermore, as exemplified by COVID-19 recently,  no domain-specific parallel data is readily available for emerging domains. 
However, the gravity of this recent calamity created a high demand for reliable translation of critical information regarding pandemic and infection prevention.
This work is part of WMT2021 Shared Task: \textit{Machine Translation using Terminologies}, where we describe Tilde MT systems that are capable of dynamic terminology integration at the time of translation. Our systems achieve up to 94\% COVID-19 term use accuracy on the test set of the EN-FR language pair without having access to any form of in-domain information during system training. 
We conclude our work with a broader discussion considering the Shared Task itself and terminology translation in MT. 
\end{abstract}
\blfootnote{*Both authors have contributed equally.}
\begin{table*}
\centering
\small
\begin{tabular}{llll}
\toprule
 & \# &\textbf{ Source Term} &\textbf{ Target Side of a Term Entry} \\
\midrule
\multirow{3}{*}{\textbf{EN-RU}} & 1 & hand-washing & \begin{otherlanguage*}{russian}мытье рук \textbar{} \textcolor{blue}{мытья рук} \textbar{} \textcolor{red}{мытья} \end{otherlanguage*} \\
 & 2 & sneeze & \begin{otherlanguage*}{russian}\textcolor{blue}{чихании} \textbar{} чихания \textbar{} \uline{чихнуть}\end{otherlanguage*} \\
 & 3 & flu epidemic & \begin{otherlanguage*}{russian}эпидемия гриппа \textbar{} \textcolor{red}{эпидемия свиного гриппа}\end{otherlanguage*} \\
\midrule
\multirow{4}{*}{\textbf{EN-FR}} & 4 & World Health Organization & \begin{tabular}[t]{@{}l@{}}Organisation mondiale de la Santé \textbar{} \textit{Organisation Mondiale de la Santé} \textbar{}\\\textit{Organisation mondiale de la santé}\end{tabular} \\
 & 5 & Coronavirus outbreak & épidémie de coronavirus \\
 & 6 & \textit{coronavirus outbreak} & \textcolor{blue}{épidémies de coronavirus} \\
\midrule
\multirow{5}{*}{\textbf{CS-DE}} & 7 & zimní paralympijské hry & \begin{tabular}[t]{@{}l@{}}\textcolor{red}{Paralympische Spiele} \textbar{} \textcolor{red}{Sommer-Paralympics} \textbar{} Paralympische Winterspiele \\ \textbar{} \textcolor{red}{Paralympische Sommerspiele} \textbar{} \textcolor{red}{Paralympics}\end{tabular} \\
 & 8 & letní paralympijské hry & \begin{tabular}[t]{@{}l@{}}\textcolor{red}{Paralympische Spiele} \textbar{} \textbf{Sommer-Paralympics} \textbar{} \textcolor{red}{Paralympische}\\ \textcolor{red}{Winterspiele} \textbar{} Paralympische Sommerspiele \textbar{} \textcolor{red}{Paralympics}\end{tabular} \\
 & 9 & narcista & Narzissmus \textbar{} \textit{Narzißmus} \textbar{} \uline{narzisstisch} \\
\bottomrule
\end{tabular}
\caption{Examples of noisy term entries in the provided term collections. Inflected forms are \textcolor{blue}{blue}, wrong translations of the source term are \textcolor{red}{red}, alternative spelling variants are \textit{italic}, other possible translations are \textbf{bold}, other terms within one term entry are \uline{underlined}.}
\label{tab:term-entry-examples}
\end{table*}

\section{Introduction}
This work is part of WMT2021 Shared Task: \textit{Machine Translation using Terminologies}, which is concerned with improving machine translation (MT) accuracy and consistency on \textit{newly developed domains} by utilising word and phrase-level terms. We describe Tilde MT systems that are capable of dynamic terminology integration at inference time. Our submissions consist of translations by terminology-enabled general-purpose MT systems for EN-RU, EN-FR, and CS-DE translation directions. Our systems are deliberately trained without consideration for the test domain to follow the spirit of the Shared Task---MT for emerging domains.
Despite term collections being noisy, our MT systems with dynamic terminology integration improve term translation accuracy proving their usefulness in dynamic adaptation for novel domains, where training-time domain adaptation methods are not feasible.

The remainder of this work describes the methods used for dynamic terminology integration (Section~\ref{sec:methods}) by describing the tasks of terminology filtering, term recognition, and dynamic terminology integration in the translation process. The bulk of Section~\ref{sec:methods} describes problems due to the low-quality term collections and terminology mismanagement and our solutions to them. We hope that the examples provided will not only illustrate the self-imposed problems by the Shared Task but also will motive reconsidering the purpose and the desired qualities of a term collection in the context of MT.
We then briefly describe the experimental setting and results in Section~\ref{sec:exp} and Section~\ref{sec:res} respectively. We conclude our work with a broader discussion considering the Shared Task and terminology translation in MT in Section~\ref{sec:concl}.

\section{Methods}\label{sec:methods}
This section describes the three tasks necessary for successful MT with terminology: terminology filtering, term recognition, and finally, integration of terminology constraints in the translation process. 

\subsection{Terminology Filtering} \label{sec:filtering}

To guarantee terminology translation correctness and consistency, which are two quality aspects of terminology translation, term collections must provide unambiguous information about the preferred translation equivalent for each source term's type (full form, short form, or acronym) when listing multiple possible translation equivalents in a target language.
Typically preparing such term collections that are useful for MT (or the translation process in general) is the task of a professional translator or a terminologist.
In this Shared Task, however, term collections sometimes contain entries that do not feature reciprocal translations of terms (e.g., see examples 1, 3, 7, and 8 in Table~\ref{tab:term-entry-examples}), have multiple translations per term (example 8 in Table~\ref{tab:term-entry-examples}), have multiple different terms merged into a single term entry (examples 2 and 9 in Table~\ref{tab:term-entry-examples}), have different spelling forms listed on the target side (examples 4 and 9 in Table~\ref{tab:term-entry-examples}), and have different inflections or spelling variants of source terms separated in different term entries (examples 5 and 6 in Table~\ref{tab:term-entry-examples}). Besides, unlike the common custom of providing terms and their translations in their dictionary forms (e.g., see examples of terms in EuroTermBank\footnote{\url{https://eurotermbank.com/}}, the InterActive Terminology for Europe\footnote{\url{https://iate.europa.eu/}}, the United Nations Terminology Database\footnote{\url{https://unterm.un.org/}}, and other authoritative term banks), the terminologies provided for the Shared Task often contain entries where the source or the target language form is already inflected (examples 1, 2, and 6 in Table~\ref{tab:term-entry-examples}).

To reduce the noise present in the provided term collections, we performed filtering by discarding:
\begin{enumerate}
    \item Term pairs that feature terms consisting of symbols other than digits, letters, apostrophes, white-spaces, and hyphen symbols. This filter allows to identify and discard expressions that do not represent terminology (e.g., full sentences, complete clauses, formulas, expressions consisting of terms and their acronyms within one term entry, etc.; see examples 1-6 and 8 in Table~\ref{tab:filtering-examples}).
    \item Term pairs where the source term is longer than the target term and the source term contains the target term as a sub-string (or vice versa). This filter is intended to discard term entries representing named entities that are written identically in both source and target languages, but for which one of the sides is incomplete (see examples 7 and 9 in Table~\ref{tab:filtering-examples}).
    \item Term pairs that represent general language (i.e., are too common). General language phrases are typically ambiguous and may require different translations based on surrounding context as well as external knowledge, which may not be available when translating. Therefore, it may be safer to allow the NMT model to handle the translation of general language phrases. We also do not want to burden the MT model too much with excessively annotated input data since longer segments are typically handled worse by NMT models than shorter segments \cite{neishi2019relation}. To identify term entries that are too general, we apply an inverse document frequency (IDF) \cite{jones1972statistical} filter \cite{pinnis-2015-dynamic-terminology}. As an example, this filter discarded from the term collections all term entries of the English term ``spread'' as it is a highly ambiguous word and according to the Collins EN-FR dictionary\footnote{\url{https://www.collinsdictionary.com/dictionary/EN-FR/spread}} it may have at least 20 distinct translations . Since the term collection features just one possible translation without any added meta-data, it is safer to not use such terms (considering also the limitations of term recognition when working with emerging domains with scarce or no parallel data).
\end{enumerate}
\begin{table*}[]
\centering
\small
\begin{tabular}{llll}
\toprule
      & \# & \textbf{Source Term}            & \textbf{Target Term    }                     \\
\midrule
\multirow{3}{*}{\textbf{EN-FR}} & 1 & shortness of breath    & essoufflé (e)                       \\
      & 2 & Do the Five            & "5 gestes" "barrière" ""            \\
      & 3 & Coronavirus (COVID-19) & coronavirus (COVID-19)              \\
\midrule
\multirow{4}{*}{\textbf{EN-RU}} & 4 & CDC                       & \begin{tabular}[t]{@{}l@{}}\begin{otherlanguage*}{russian}центры по контролю и профилактике \end{otherlanguage*}\\\begin{otherlanguage*}{russian}заболеваний США \end{otherlanguage*}(CDC)\end{tabular} \\
      & 5 & active cases           & \begin{otherlanguage*}{russian}активные случаи [заболевания] \end{otherlanguage*}       \\
      & 6 & contagious             & \begin{otherlanguage*}{russian}заразный: передающийся при контакте \end{otherlanguage*}\\
      \midrule
\multirow{3}{*}{\textbf{CS-DE}} & 7 & Lions Clubs International & Lions Clubs                                             \\
      & 8 & VIII. hlavový nerv     & Nervus vestibulocochlearis          \\
      & 9 & Eli Lilly              & Eli Lilly and Company        \\      
\bottomrule
\end{tabular}
\caption{Examples of terms removed by basic filtering.}
\label{tab:filtering-examples}
\end{table*}
Besides filtering out the noisy term entries, the type and the quality of the term collections provided in the Shared Task also require selecting one among potentially many term translation equivalents. As noted before, this is typically done by a human, possibly, a domain expert. Nevertheless, we opt for two different strategies. If more than one translation equivalent is provided, it is fair to assume that they are all equally applicable. Thus we propose to select the first translation equivalent in the list. We refer to it as \textbf{1\textsuperscript{st} Trg Term} term selection strategy. After analyzing the term collections, however, we conclude that it is not the case that all translation equivalents provided in the Shared Task are of equal quality. Therefore, we employ a statistical word alignment-based strategy to select the translation equivalent with the highest alignment score. To compute word alignments we use \textit{eflomal}\footnote{\url{https://github.com/robertostling/eflomal}} \cite{Ostling2016efmaral}. We refer to it as \textbf{Alignment-based} term selection strategy.

Table~\ref{tab:term-select-examples} gives examples of terms selected by either of the term selection strategy. Examples illustrate that some translation equivalents are of equal quality (examples 1, 3, 6, and 7). However, selecting the first translation equivalent can sometimes give long (examples 6 and 8) or inadequate (example~4) translation equivalents. The Alignment-based term selection strategy also tends to select translation equivalents that are dictionary forms (example~2) instead of inflections.

\subsection{Term Recognition}
Having a term collection, the next task in the MT workflow is term recognition in a running text. Term recognition depending on the morphological typology of the source language and the nature of the domain can prove to be a complex task. Recognition involves term identification in its surface form, which for morphologically complex languages may be hindered by many surface forms a single word can take or by the level of form ambiguity in the case of morphologically impoverished languages \cite{bergmanis2018context}. To overcome issues posed by the morphology of the natural language, one can use one of the many off-the-shelf morphological taggers to obtain contextually correct part-of-speech and lemma pairs for each token and perform term recognition on lemmatized collections and texts. We, however, opt for an alternative, a more rudimentary method utilizing language-specific stemmers to normalise the surface forms and do the term recognition on stemmed running text and term collections \cite{pinnis-2015-dynamic-terminology, pinnis2015terminology}.
We opt for the stemmer-based approach because, in the production setting, stemming is faster than morphological tagging and has broader coverage for low-resource languages. Besides, to take full advantage of the morpho-syntactic information provided by morphological taggers, similar information must be provided by the term collection. However, as term collections of this Shared Task exemplify, expecting any meta-data is naive.

Last but not least, recognised forms must be word-sense disambiguated if more than one translation sense (i.e., term entry per source-side lexical form) is available. Word-sense disambiguation tools typically are lexicalized classifiers that are trained using large amounts of parallel data. However, the spirit of this Shared Task is MT using terminology for emerging domains where "\textit{parallel data are hard to come by}"\footnote{\url{http://www.statmt.org/wmt21/terminology-task.html}}. Thus we skip word-sense disambiguation and use just one word-sense per word form. 

\subsection{Integration of Terminology Constraints}
In the day-to-day work of professional translators, terminologies are glossaries containing source language terms and their corresponding target language translations in their dictionary forms. Some of the previous work on terminology translation assumes that term entries are given in forms already inflected as required by the target morpho-syntactic context. Thus, such work focuses either on morphologically impoverished languages or is concerned with terminology translation in unrealistic scenarios. Either way, such methods are not relevant for the languages of the Shared Task because all of the target languages, with the exception of Chinese, are to some degree inflective languages. 

There is, however, another body of work that addresses translation with terminology while accounting for morphological complexity of the target language \cite{exel-etal-2020-terminology,niehues2021continuous,bergmanis2021facilitating}. 
We base our submission on \newcite{bergmanis2021facilitating} and employ target lemma annotations (TLA) to augment MT training data. An example of a sentence fragment annotated with TLA is
"\textit{infections}\textbf{|s} \begin{otherlanguage*}{russian}\textit{инфекция}\end{otherlanguage*}\textbf{|t} \textit{result}|w \textit{in}\textbf{|w} \textit{mild}\textbf{|w} \textit{symptoms}\textbf{|w}", where \textbf{|s}, \textbf{|t}, \textbf{|w} are factors indicating whether token is a source language term, a lemma (the dictionary form) of a target language term, or an ordinary source language word respectively. 
Systems trained on such data are equipped with a mechanism for passing soft terminology constraints at inference time. An essential property of MT systems trained using TLA is that they learn not just to copy but also to inflect the provided terminology constraints according to the target morphosyntactic context. Therefore, the translation of the sentence above, for example, "\begin{otherlanguage*}{russian}инфекций приводят к легким симптомам\end{otherlanguage*}", contains the plural noun "\begin{otherlanguage*}{russian}инфекций\end{otherlanguage*}" and not just the annotated singular form "\begin{otherlanguage*}{russian}инфекция\end{otherlanguage*}".
\begin{table*}[]
\small
\centering
\begin{tabular}{lclll}
\toprule
                     & \#  & \textbf{Source Term }          & \textbf{1\textsuperscript{st} Trg Term }                  & \textbf{Alignment-based}       \\
\midrule
\multirow{3}{*}{\textbf{EN-FR}}               & 1 & disease outbreak      & apparition de maladie          & épidémie              \\
                      & 2 & epidemiologist        & épidémiologistes               & épidémiologiste       \\
                      & 3 & wash your hands       & lavez-vous les mains           & laver les mains       \\
\midrule
\multirow{8}{*}{\textbf{EN-RU}} &
  4 &Center for Disease Control &\begin{otherlanguage*}{russian}\begin{tabular}[t]{@{}l@{}}центры по контролю\\и профилактике\\заболеваний США \end{tabular} \end{otherlanguage*}&\begin{otherlanguage*}{russian}\begin{tabular}[t]{@{}l@{}}Центр контроля\\ и профилактики \\заболеваний \end{tabular} \end{otherlanguage*}\\
  & 5                     & COVID-19 crisis       & \begin{otherlanguage*}{russian}кризиса, связанного с COVID-19 \end{otherlanguage*}& \begin{otherlanguage*}{russian}кризис из-за COVID-19\end{otherlanguage*}\\
 
  & 6 & health crisis &   \begin{otherlanguage*}{russian}\begin{tabular}[t]{@{}l@{}}кризисной ситуации,\\которая сложилась сегодня\\в сфере общественного\\здравоохранения \end{tabular}\end{otherlanguage*} &  \begin{otherlanguage*}{russian}кризис здравоохранения \end{otherlanguage*}\\
  \midrule
\multirow{4}{*}{\textbf{CS-DE}} &6 &  benzoylperoxid        & Dibenzoylperoxid               & Benzoylperoxid        \\
                       & 7 & alternativní medicína & Alternativmedizin              & Alternative Medizin   \\
 & 8 & AV ČR  & \begin{tabular}[t]{@{}l@{}}Akademie der Wissenschaften\\ der Tschechischen Republik\end{tabular} &
  AV ČR \\
  \bottomrule
\end{tabular}
\caption{Examples of terms selected by either \textbf{1\textsuperscript{st} Trg Term} or \textbf{Alignment-based} term selection strategy for term entries with multiple translation equivalents.}
\label{tab:term-select-examples}
\end{table*}

\section{Experiments}\label{sec:exp}
\paragraph{Data.}
We use all parallel data provided for the Shared Task for training, except for development data which we use to choose the best model for final submission. Although back-translated monolingual data could, in theory, improve the overall translation quality, we do not use it to train our systems because typically, the monolingual target data is selected based on its similarity with the target domain data. However, the scenario proposed for the Shared Task assumes that the domain is novel; thus, we aim to explore the merits of terminology translation and do not look for extra synthetic target domain data.

\paragraph{MT Model and Training.}
For system training, we use the \textit{Marian toolkit} \cite{mariannmt} because of its factored model functionality developed within the scope of the \textit{User-Focused Marian project} \footnote{\url{https://marian-project.eu/}}\footnote{\url{https://github.com/marian-cef/marian-examples/blob/forced-translation/forced-translation/docs/Experiments.md}}.
In this Shared Task, we train standard MT systems that mostly follow the Transformer \cite{vaswani2017attention} \textit{base} model configuration. The only deviations from the standard configuration are 1) the use of source-side factors (we use factor embeddings of dimensionality 8 and concatenate them with word embeddings), 2) increased \textit{--optimizer-dealy} (from 16 to 24), and 3) an increased maximum sequence length (from 128 to 196 tokens). These changes are necessary purely for TLA support during training and inference: increased sequence length accounts for longer input sequences due to TLA and terminology constraints, while increased optimizer delay compensates for fewer sentences fitting in workspace memory-based batch due to their increased maximum length.
\begin{table*}
\centering
\begin{tabular}{lccccc}
\toprule
\textbf{EN-FR     }             & \textbf{BLEU}                                     & \begin{tabular}[c]{@{}l@{}}\textbf{Accuracy}\end{tabular} & \textbf{Window 2}                                  & \textbf{Window 3}                                  & \textbf{1 - TERm~}                                  \\
\midrule
Baseline               & {\cellcolor[rgb]{0.549,0.796,0.518}}44.5 & {\cellcolor[rgb]{1,0.937,0.612}}0.885                          & {\cellcolor[rgb]{0.698,0.843,0.549}}0.286 & {\cellcolor[rgb]{0.827,0.886,0.576}}0.278 & {\cellcolor[rgb]{0.51,0.784,0.51}}0.575    \\
1\textsuperscript{st} Trg Term    & {\cellcolor[rgb]{1,0.937,0.612}}44.2     & {\cellcolor[rgb]{0.549,0.796,0.518}}0.922                      & {\cellcolor[rgb]{0.812,0.878,0.573}}0.284 & {\cellcolor[rgb]{0.8,0.875,0.573}}0.278   & {\cellcolor[rgb]{1,0.937,0.612}}0.572      \\
1\textsuperscript{st} Trg Term, IDF$\geq$5    & {\cellcolor[rgb]{0.78,0.871,0.569}}44.3  & {\cellcolor[rgb]{0.761,0.863,0.565}}0.905                      & {\cellcolor[rgb]{1,0.937,0.612}}0.281     & {\cellcolor[rgb]{1,0.937,0.612}}0.275     & {\cellcolor[rgb]{0.765,0.867,0.565}}0.574  \\
1\textsuperscript{st} Trg Term, IDF$\geq$7    & {\cellcolor[rgb]{0.714,0.847,0.553}}44.4 & {\cellcolor[rgb]{0.937,0.918,0.6}}0.890                        & {\cellcolor[rgb]{0.784,0.871,0.569}}0.285 & {\cellcolor[rgb]{0.71,0.847,0.553}}0.280  & {\cellcolor[rgb]{0.745,0.859,0.561}}0.574  \\
\textbf{Alignment-based} & {\cellcolor[rgb]{0.42,0.757,0.49}}44.6   & {\cellcolor[rgb]{0.388,0.745,0.482}}\textbf{0.936 }                     & {\cellcolor[rgb]{0.388,0.745,0.482}}0.291 & {\cellcolor[rgb]{0.388,0.745,0.482}}0.285 & {\cellcolor[rgb]{0.388,0.745,0.482}}0.576  \\
Alignment-based, IDF$\geq$5 & {\cellcolor[rgb]{0.388,0.745,0.482}}44.6 & {\cellcolor[rgb]{0.604,0.816,0.529}}0.918                      & {\cellcolor[rgb]{0.624,0.82,0.533}}0.287  & {\cellcolor[rgb]{0.639,0.827,0.537}}0.281 & {\cellcolor[rgb]{0.435,0.761,0.494}}0.576  \\
Alignment-based, IDF$\geq$7 & {\cellcolor[rgb]{0.702,0.843,0.549}}44.4 & {\cellcolor[rgb]{0.871,0.898,0.584}}0.896                      & {\cellcolor[rgb]{0.827,0.886,0.576}}0.284 & {\cellcolor[rgb]{0.796,0.875,0.569}}0.278 & {\cellcolor[rgb]{0.745,0.859,0.561}}0.574  \\
\midrule
\textbf{EN-RU    }              & \textbf{BLEU}                                     & \begin{tabular}[c]{@{}l@{}}\textbf{Accuracy}\end{tabular} & \textbf{Window 2}                                  & \textbf{Window 3      }                            &\textbf{ 1 - TERm~}                                  \\
\midrule
Baseline               & {\cellcolor[rgb]{0.663,0.831,0.541}}24.9 & {\cellcolor[rgb]{0.933,0.918,0.6}}0.760                        & {\cellcolor[rgb]{0.996,0.937,0.612}}0.163 & {\cellcolor[rgb]{1,0.937,0.612}}0.164     & {\cellcolor[rgb]{0.718,0.851,0.553}}0.398  \\
1\textsuperscript{st} Trg Term    & {\cellcolor[rgb]{1,0.937,0.612}}24.6     & {\cellcolor[rgb]{1,0.937,0.612}}0.751                          & {\cellcolor[rgb]{1,0.937,0.612}}0.163     & {\cellcolor[rgb]{0.988,0.937,0.612}}0.164 & {\cellcolor[rgb]{1,0.937,0.612}}0.394      \\
1\textsuperscript{st} Trg Term, IDF$\geq$5    & {\cellcolor[rgb]{0.588,0.808,0.525}}25.0 & {\cellcolor[rgb]{0.522,0.788,0.514}}0.815                      & {\cellcolor[rgb]{0.424,0.757,0.49}}0.176  & {\cellcolor[rgb]{0.451,0.765,0.498}}0.176 & {\cellcolor[rgb]{0.471,0.773,0.502}}0.401  \\
1\textsuperscript{st} Trg Term, IDF$\geq$7    & {\cellcolor[rgb]{0.475,0.773,0.502}}25.1 & {\cellcolor[rgb]{0.639,0.824,0.537}}0.800                      & {\cellcolor[rgb]{0.51,0.784,0.51}}0.174   & {\cellcolor[rgb]{0.498,0.78,0.506}}0.175  & {\cellcolor[rgb]{0.482,0.776,0.502}}0.401  \\
\textbf{Alignment-based} & {\cellcolor[rgb]{0.561,0.8,0.522}}25.0   & {\cellcolor[rgb]{0.388,0.745,0.482}}\textbf{0.833}                      & {\cellcolor[rgb]{0.494,0.78,0.506}}0.174  & {\cellcolor[rgb]{0.486,0.776,0.506}}0.175 & {\cellcolor[rgb]{0.388,0.745,0.482}}0.403  \\
Alignment-based, IDF$\geq$5 & {\cellcolor[rgb]{0.455,0.769,0.498}}25.1 & {\cellcolor[rgb]{0.482,0.776,0.506}}0.821                      & {\cellcolor[rgb]{0.388,0.745,0.482}}0.176 & {\cellcolor[rgb]{0.388,0.745,0.482}}0.177 & {\cellcolor[rgb]{0.416,0.757,0.49}}0.402   \\
Alignment-based, IDF$\geq$7 & {\cellcolor[rgb]{0.388,0.745,0.482}}25.2 & {\cellcolor[rgb]{0.694,0.843,0.549}}0.792                      & {\cellcolor[rgb]{0.639,0.827,0.537}}0.171 & {\cellcolor[rgb]{0.627,0.82,0.533}}0.172  & {\cellcolor[rgb]{0.471,0.773,0.502}}0.401  \\
\midrule
\textbf{CS-DE }                 & \textbf{BLEU }                                    & \begin{tabular}[c]{@{}l@{}}\textbf{ Accuracy}\end{tabular} &\textbf{ Window 2}                                  &\textbf{ Window 3  }                                & \textbf{1 - TERm~}                                  \\
\midrule
Baseline               & {\cellcolor[rgb]{0.388,0.745,0.482}}29.9 & {\cellcolor[rgb]{1,0.937,0.612}}0.824                          & {\cellcolor[rgb]{0.388,0.745,0.482}}0.376 & {\cellcolor[rgb]{0.388,0.745,0.482}}0.368 & {\cellcolor[rgb]{0.388,0.745,0.482}}0.390  \\
1\textsuperscript{st} Trg Term    & {\cellcolor[rgb]{1,0.937,0.612}}26.3     & {\cellcolor[rgb]{0.424,0.757,0.49}}0.851                       & {\cellcolor[rgb]{1,0.937,0.612}}0.338     & {\cellcolor[rgb]{1,0.937,0.612}}0.328     & {\cellcolor[rgb]{1,0.937,0.612}}0.340      \\
1\textsuperscript{st} Trg Term, IDF$\geq$5    & {\cellcolor[rgb]{0.471,0.773,0.502}}29.5 & {\cellcolor[rgb]{0.396,0.749,0.486}}0.852                      & {\cellcolor[rgb]{0.471,0.773,0.502}}0.371 & {\cellcolor[rgb]{0.494,0.78,0.506}}0.361  & {\cellcolor[rgb]{0.541,0.792,0.518}}0.378  \\
1\textsuperscript{st} Trg Term, IDF$\geq$7    & {\cellcolor[rgb]{0.439,0.761,0.494}}29.6 & {\cellcolor[rgb]{0.71,0.847,0.553}}0.837                       & {\cellcolor[rgb]{0.439,0.761,0.494}}0.373 & {\cellcolor[rgb]{0.447,0.765,0.498}}0.364 & {\cellcolor[rgb]{0.502,0.784,0.51}}0.381   \\
Alignment-based & {\cellcolor[rgb]{0.925,0.914,0.596}}26.8 & {\cellcolor[rgb]{0.459,0.769,0.498}}0.849                      & {\cellcolor[rgb]{0.878,0.902,0.588}}0.346 & {\cellcolor[rgb]{0.871,0.898,0.584}}0.337 & {\cellcolor[rgb]{0.906,0.91,0.592}}0.348   \\
\textbf{Alignment-based, IDF$\geq$5} & {\cellcolor[rgb]{0.455,0.769,0.498}}29.6 & {\cellcolor[rgb]{0.388,0.745,0.482}}\textbf{0.853}                      & {\cellcolor[rgb]{0.435,0.761,0.494}}0.373 & {\cellcolor[rgb]{0.451,0.765,0.498}}0.364 & {\cellcolor[rgb]{0.443,0.765,0.494}}0.386  \\
Alignment-based, IDF$\geq$7 & {\cellcolor[rgb]{0.424,0.757,0.49}}29.7  & {\cellcolor[rgb]{0.71,0.847,0.553}}0.837                       & {\cellcolor[rgb]{0.416,0.757,0.49}}0.375  & {\cellcolor[rgb]{0.416,0.757,0.49}}0.366  & {\cellcolor[rgb]{0.412,0.753,0.49}}0.389  \\
\bottomrule
\end{tabular}
\caption{Development set results for baseline MT systems and systems using terminology integration. MT systems using terminology integration are named after the method used for terminology filtering. The  numerically highest scores according the Lemmatized Exact-Match Accuracy (Accuracy) in data setting are \textbf{bold}. For detailed description of other metrics consult \cite{alam2021evaluation}.}
\label{tab:results}
\end{table*}

\section{Results}\label{sec:res}

We trained one NMT system per translation direction and evaluated translation quality on the development sets using the terminology translation evaluation tool provided by the Shared Task\footnote{\url{https://github.com/mahfuzibnalam/terminology_evaluation}} \cite{alam2021evaluation}. We compare the baseline translation scenario where no terms are annotated in the source text with improved scenarios where terms are annotated using term collections acquired using the different filtering and term translation equivalent selection strategies.

When analysing the lemmatized exact match accuracy, we must bear in mind that the evaluation data similarly to the term collections feature term entries 1) with more than one allowed synonymous translation equivalent (not counting different inflected forms), and 2) where different terms are merged into one entry (see examples in Section~\ref{sec:filtering}). This consequently means that the evaluation procedure 1) allows terminological ambiguity on the target side, 2) does not allow analysing terminology translation consistency, and 3) may depict a rough estimate of terminology translation accuracy. Therefore, we believe that the lemmatized exact match accuracy results should be analysed with a grain of salt.

That being said, the results in Table~\ref{tab:results} show that the metric improves when using a term collection in all but one experiment. The fact that in overall the Alignment-based term collections show better overall translation results (in terms of BLEU) and also allow reaching the highest terminology translation accuracy results shows that relying on the first translation equivalent in a term entry is not a good idea.

We also see that the overall terminology translation quality is already relatively high for the baseline systems ranging from 76\% for EN-RU till 88.5\% for EN-FR. This makes us wonder whether the evaluated domain can be considered emerging as it features few novel terms and the majority are well handled by the baseline systems. To investigate further, we analysed whether the bilingual terminology that can be found in the development sets is also present in the training data of the NMT systems. We found that for CS-DE, 97.9\% and 92.5\% of such (unique) bilingual terms are featured at least once or 10 times in the training data respectively. The numbers are even higher if we analyse running terms (tokens) -- 99.8\% and 98.7\% respectively. Since the terminology for EN-RU was human-created and not extracted from parallel data, it shows slightly lower results when analysing unique terms -- 93.5\% and 88.3\% respectively. However, the situation is similar to CS-DE when analysing running terms -- 99.1\% and 97.8\% of bilingual terms found in the EN-RU development data are also found in the training data at least once and 10 times respectively. Based on these findings, we believe that the validation data does not depict an emerging domain and does not help analysing terminology translation quality for emerging domains.

When analysing the overall translation quality (in terms of BLEU), we see that term filtering using the IDF-based filter is crucial when relying on very noisy and automatically acquired term collections (as was the case with the CS-DE term collection). The results show that translation quality drops by 3 BLEU points when using the unfiltered term collections. This shows that too general (and ambiguous) terminology can be harmful and lower translation quality. The overall translation quality change is marginal for the translation directions that featured human-created term collections (EN-FR and EN-RU), however we do see an increase in terminology translation accuracy.

Our \textbf{final submission} consists of machine translations of Shared Task test sets provided by general-purpose MT systems that use dynamic terminology integration using TLA \cite{bergmanis2021facilitating}.  To translate our final submissions, term collections are filtered by basic filters (see Section~\ref{sec:filtering}) for EN-FR and EN-RU language pairs, while for CS-DE language pair, we also use IDF>5 filtering. We use the statistical word alignment term selection strategy for term entries with multiple translation equivalents for all language pairs. The development set results for the corresponding systems are marked bold in Table~\ref{tab:results}.

\section{Discussion}\label{sec:concl}
\paragraph{Shared Task.}
Results of automatic metrics show that our baseline systems are already well equipped to translate the development and test sets regardless of their seemingly novel domain. Indeed -- we found no statistically significant differences in scores measuring general translation quality between the baseline systems and systems with terminology integration. Preliminary test results suggest a similar pattern in other submissions (c.f. results of submissions by Prompt). The only seemingly meaningful differences are in metrics specifically targeting terminology integration. These results are in stark contrast with previous work \cite{exel-etal-2020-terminology,niehues2021continuous,bergmanis2021facilitating} which report significant improvements not only on terminology use targeted metrics but also on metrics measuring general translation quality. This disparity suggests that test data is \textit{not} from an emerging or novel domain (at least as far MT systems trained on the training data provided are concerned). Considering this shortcoming, together with the visibility of WMT Shared Tasks, these results pose a risk of misrepresenting the problem the Shared Task was set out to research. The outcome might be unintended downplaying of the role of terminology translation for technical domains, which could lead to diminishing interest in terminology translation from the MT research community.
\vspace{-5px}
\paragraph{Term Collections in MT.}
Tables~\ref{tab:term-entry-examples}, \ref{tab:filtering-examples}, and \ref{tab:term-select-examples} of  Section~\ref{sec:methods} provide numerous examples of problems present in the provided term collections. We
believe that these examples illustrate the understanding of the purpose and desired qualities of a term collection not just of those individuals involved in preparing term collections for the Shared Task but also to a broader community of translation professionals. Many of the problematic examples suggest that the shift from human-readable to machine-readable term collections is not there yet, or that it has happened rather formally by merely reformatting the for-human-made term collections into neater TSV-formatted files. While having TSV-formatted files helps for the file to be machine-readable, it does not make the content machine-usable.  The standards for-machine-made term collections have to be higher than those made for humans. At least as long as there is no sophisticated intelligence in the MT workflow that is on par with humans to recover from the irregularities and noise present in the term collections typically made for humans. Likewise, the encyclopedia-style entries explaining a concisely coined concept of the source language using a whole sentence to define it in the target language are still present in for-human-made term collections, but they are of no use for current MT systems.

If translation with terminology is supposed to improve MT for novel domains, the term collections, being the supposed source of the expected improvement, have to be of a higher quality than the MT systems they are intended to improve.
\vspace{-5px}

\section*{Acknowledgements}
This research has been supported by the User-focused Marian project which is  co-financed by the European Union's Connecting Europe Facility programme  (Action number: 2019-EU-IA-0045).
\vspace{-5px}

\bibliographystyle{acl_natbib}
\bibliography{custom}
\appendix

\end{document}